\documentclass[10pt,twocolumn,letterpaper]{article}

\usepackage{cvpr}
\usepackage{times}
\usepackage{epsfig}
\usepackage{graphicx}
\usepackage{amsmath}
\usepackage{amssymb}

\usepackage{subfigure}
\usepackage{mathtools}
\usepackage{multirow}

\usepackage[breaklinks=true,bookmarks=false]{hyperref}

\cvprfinalcopy % *** Uncomment this line for the final submission

\def\cvprPaperID{8071} % *** Enter the CVPR Paper ID here

\begin{document}

%%%%%%%%% TITLE
\title{Bi-Semantic Reconstructing Generative Network for Zero-shot Learning}

\author{Shibing Xu\\
Xidian University,China\\
{\tt\small sbxu@stu.xidian.edu.cn}
\and Zishu Gao\\
Beijing University Of Chemical Technology,China\\
{\tt\small zishugaoo@163.com}
\and Guojun Xie\\
Xidian University,China\\
{\tt\small sbxu@stu.xidian.edu.cn}
}

\author{Shibing Xu$^1$, Zishu Gao$^2$, Guojun Xie$^1$\\
$^1$Xidian University,China;$^2$Beijing University Of Chemical Technology,China\\
{\tt\small sbxu@stu.xidian.edu.cn}
}

\maketitle
%\thispagestyle{empty}

%%%%%%%%% ABSTRACT
\begin{abstract}
		Many recent methods of zero-shot learning (ZSL) attempt to utilize generative model to generate the unseen visual samples from semantic descriptions and random noise. Therefore, the ZSL problem becomes a traditional supervised classification problem. However, most of the existing methods based on the generative model only focus on the quality of synthesized samples at the training stage, and ignore the importance of the zero-shot recognition stage. In this paper, we consider both the above two points and propose a novel approach. Specially, we select the Generative Adversarial Network (GAN) as our generative model. In order to improve the quality of synthesized samples, considering the internal relation of the semantic description in the semantic space as well as the fact that the seen and unseen visual information belong to different domains, we propose a bi-semantic reconstructing (BSR) component which contain two different semantic reconstructing regressors to lead the training of GAN. Since the semantic descriptions are available during the training stage, to further improve the ability of classifier, we combine the visual samples and semantic descriptions to train a classifier. At the recognition stage, we naturally utilize the BSR component to transfer the visual features and semantic descriptions, and concatenate them for classification. Experimental results show that our method outperforms the state of the art on several ZSL benchmark datasets with significant improvements\footnote {Our source code is available in \url {https://github.com/dfwehs/bsrgan}.}.
\end{abstract}

%%%%%%%%% BODY TEXT
\section{Introduction}

\begin{figure}[t]
\begin{center}
	\subfigure[]{
		\centering
		\includegraphics[width=0.8\linewidth]{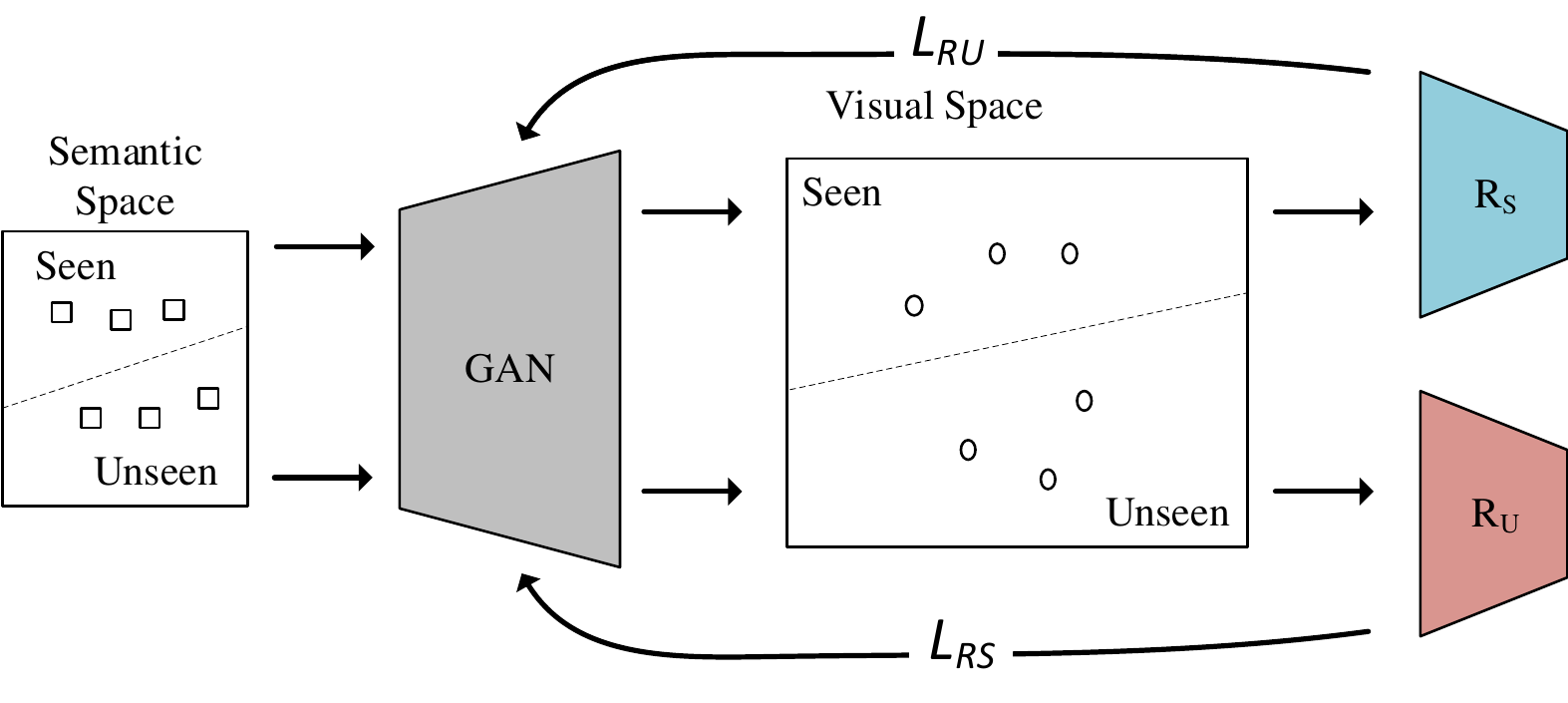}
	}
	\subfigure[]{
		\centering
		\includegraphics[width=0.8\linewidth]{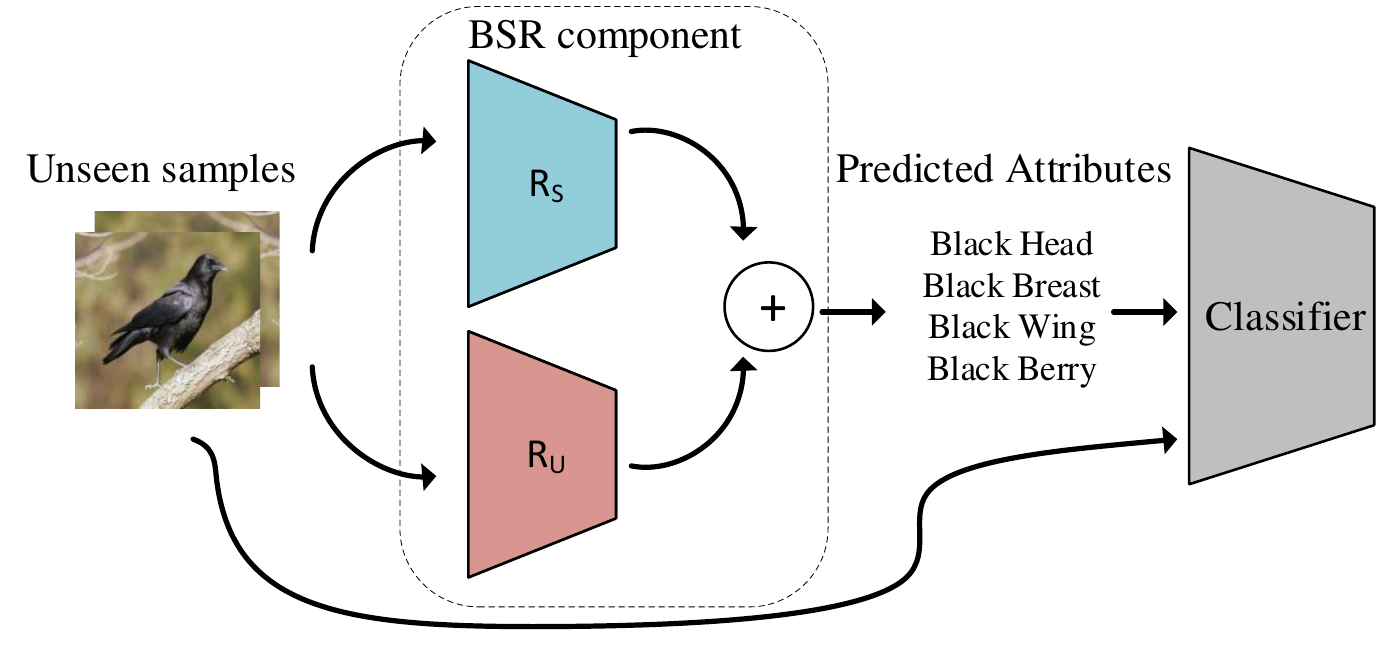}
	}
\end{center}
   \caption{Illustration of our proposed model. (a) The generative model is constrained by the seen and unseen reconstruction losses calculated by two regressor respectively. (b) Test visual samples are concatenated with its reconstructed semantic description as the inputs of classifier for classification at recognition stage.}
\label{fig:f1}
\end{figure}

	Over past several years, there has been great success in image classification task achieved by deep learning model. The deep learning model heavily relies on fully supervised training at large scale dataset of annotated samples. However, due to the existence of the long-tail distribution effect~\cite{LONGTAIL}, there are tremendous objects in the real life without or in lack of real data, which makes the traditional supervised learning impracticable. Targeting on tackling such a problem to recognize unseen objects, zero-shot learning (ZSL)~\cite{ImageNet} has been widely researched recently. 

	The goal of zero-shot learning is to learn a classifier from the set of seen classes with labeled samples, and then test it at the set of unseen classes without labeled training samples, where seen and unseen classes is totally disjoint. In order to relate the visual samples from seen and unseen classes, they are commonly represented as semantic descriptions. There are several ways to represent semantic descriptions, such as by predefined attributes~\cite{Attribute, DAP}, which describe well-known common characteristics of objects, or by text features extract from wiki~\cite{GAZSL}. With the help of semantic descriptions, many previous methods~\cite{DAP, DEM, SJE, SYNC, DEVISE} attempt to learn a mapping function to transfer all of the test visual samples into the same semantic space. In this way, the transformed semantic description can be graded and sorted with the semantic description of all test classes, and the categories of test samples can be predicted. However, in the face of the generalized zero-shot learning (GZSL)~\cite{Zongshu1, Zongshu2} problem which need the ability to classify images from seen and unseen classes at test stage, these methods suffer from the domain shift problem~\cite{Dominshift} and perform unsatisfactory due to the distributional difference between seen and unseen classes. To solve this problem, some recent works~\cite{GAZSL, FCLA, LISGAN, DGAN, GMGAN, G1} utilize generative model to generate samples of unseen classes, and transfer the zero-shot learning problem to fully traditional supervised classification problem. Similarly, we want to utilize the ability of generative model to generate samples of unseen classes from semantic descriptions and random noise sampling from Gaussian distribution. Assuming that the accuracy of zero-shot learning depends on the quality of synthesized samples and classification ability classifier, we focus on the following two problems: (1) How to improve the separability and the relevance to semantic descriptions of synthesized samples? (2)How to maximize the ability of classifier at zero-shot recognition stage? 

	In this paper, we utilize Generative Adversarial Network (GAN) to generate samples of unseen classes, and apply semantic descriptions to control the classes of synthesized samples. Inspired by ~\cite{DGAN, SAE}, we attempt to use dual learning to improve the quality of synthesized samples. The seen samples and unseen samples belong to different domains, and only the seen samples are available at train stage, the synthesized samples would close to the seen samples, and debase the performance of GZSL. Therefore, as shown in Fig.~\ref{fig:f1} (a), we proposed a bi-semantic reconstructing (BSR) component which use two different semantic reconstructing regressors to reconstruct generated samples into semantic descriptions, and utilize the reconstruction losses to constrain generative model and improve the quality of synthesized samples. 

	In order to improve the ability of classifier, we propose a novel recognition method which called visual semantic recognition (VSR) as shown in Fig.~\ref{fig:f1}(b). Considering the seen and unseen descriptions are available on train stage, we attempt to combine semantic descriptions with the visual samples as the inputs of classifier to improve its recognition ability at zero-shot recognition stage. In this way, our problem is turned into how to transfer the visual samples into semantic descriptions. We naturally have an idea to use BSR component to do this transfer. In this paper, we adopt a simple weighted addition operation to combine the results of the two regressors in BSR component to obtain the reconstructed semantic descriptions, and finally combine the semantic descriptions with the visual samples for classification tasks.  

	The contributions of our work are summarized as follows: 

	(1) We propose a novel generative model for zero-shot learning which takes the advantage of generative adversarial network. To improve the quality of synthesized samples, considering the domain difference between seen and unseen classes, we proposed BSR component with two semantic reconstructing regressors to calculate reconstruction losses of seen synthesized samples and unseen synthesized samples to constrain the generative model. 

	(2) To improve the ability of classifier, we proposed a new recognition method VSR by combining the visual samples and semantic descriptions as the inputs of it with the help of BSR component. 

	(3) Extensive experiments were carried out on several widely-used datasets, verifying that the methods we proposed in this paper are superior to the existing methods with great improvements.

\begin{figure*}
\begin{center}
%\fbox{\rule{0pt}{2in} \rule{.9\linewidth}{0pt}}
\includegraphics[width=0.9\linewidth]{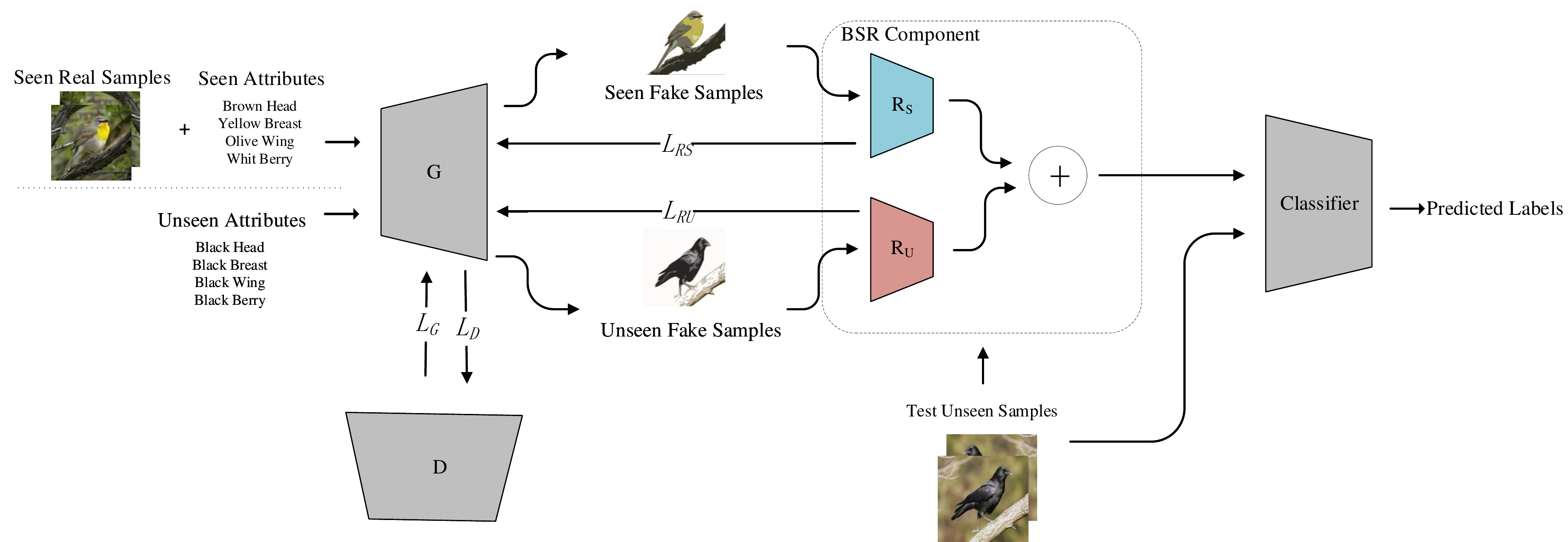}
\end{center}
   \caption{Detailed illustration of our proposed methods. We train the conditional Wasserstein GAN to generate samples of unseen classes, and utilize the reconstruction losses calculated with the help of the BSR component of seen and unseen classes to improve the quality of synthesized samples. At zero-shot recognition stage, the visual sample and its real semantic description is concatenated as the train set to train a classifier. Then, the classifier predicts the labels of test set from the visual samples and its reconstructed semantic descriptions by the BSR component.}
\label{fig:f2}
\end{figure*}

%-------------------------------------------------------------------------
\section{Related Work}
\subsection{zero-shot learning}
	Zero-shot learning (ZSL) was first proposed by~\cite{GAZSL}, which aims to leverage seen visual samples and semantic descriptions to train a classifier with good generalization ability to recognize unseen object by only giving some semantic descriptions. Over years in the past, a large numbers of ZSL methods have been proposed. Many methods attempt to learn a direct mapping function to project semantic descriptions and visual samples into the same embedding space. In this way, the classes label of test classes can be predicted by ranking the similarly score between semantic descriptions of all test classes and transferred from instance. Some works~\cite{DAP, DEM, SJE, SYNC, DEVISE} attempt to learn a mapping between visual space and semantic space, then transfer test visual samples into semantic space. In contrast, some methods~\cite{VS1, VS2, RAC} learn a semantic visual mapping to transfer the semantic descriptions into visual space for classification. There are also some works~\cite{SYNC, VSE1, VSE2} which attempt to learn a intermediate space share by visual samples and semantic descriptions, and project both of them into the intermediate space share for predicting labels. 

	Affected by domain shift problem, those methods that directly learn a mapping function perform unsatisfactory on generalized zero-shot leaning (GZSL)~\cite{GZSL1}, which joint seen samples into test set at recognition stage. Recently, the use of generative model has gained extensive attention. ~\cite{DGAN} proposes a new generative model combine the Generate Adversarial Network and Variational Auto-Encoder, and utilizes a regressor to rebuild the semantic descriptions of synthesized samples. ~\cite{LISGAN} utilizes sour samples to improve quality of synthesized samples. 

	Among these works, ~\cite{DGAN} is the closest one to our work, which applies a regressor to calculate reconstruction loss. However, there are two major differences from ours. First, we utilize different regressors to reconstruct seen and unseen samples respectively for the domain difference between seen and unseen classes. Second, in order to improve the classification accuracy, we propose a novel recognition method, which combines the visual samples and semantic descriptions to train a classifier, and utilizes test visual samples and the reconstructed semantic descriptions at test stage to predict labels of test samples. Experiment results show that our methods made great improvements than this work.

\subsection{Generative Adversarial Network}
	A typically Generative Adversarial Network (GAN) ~\cite{GAN} consists of a generator and a discriminator, and two components trained in an adversarial manner. The generator attempts to transfer a standard distribution to a synthesized distribution to approximate real distribution, and the discriminator attempts to learn how to measure the distance between the real distribution and the synthesized distribution. 

	GAN has shown great performance on generating realistic images~\cite{GANP1, GANP2, GANP3, DCGAN, G2}. However, GAN is known of its instability. In order to solve this problem and improve the quality of generated data, a lot of study has been devoted.~\cite{WGANA} shows that the use of the Jenson-Shannon divergence to lead the discriminator to measure the distance between distributions result in the training instability.~\cite{WGANCP} proposes WGAN, which optimizes an efficient approximation of Wasserstein distance via weight clipping to enforce the 1-Lipschitz constraint on discriminator. Since weight clipping would cause the vanishing and exploding gradient problems,~\cite{WGANGP} applies gradient penalty to replace weight clipping to enforce the Lipschitz constraint. In addition,~\cite{CGAN} proposes to utilize conditional messages to incorporate the other information into the generator to synthesize specified samples. In our work, we deploy WGAN, which is constrained by semantic information, to generate data.

%-------------------------------------------------------------------------
\section{Methodology}
\subsection{Problem Define}
	  In this paper, we study both ZSL and GZSL. Let $S=\{(x, y, c)|x \in X_s, y \in Y_s, c \in C_s\}$  be the training set, where $x$ is a visual sample, \emph{i.e.} a visual feature extracted through a pre-trained neural network, and $X_s$ represents the collection of all visual samples from seen classes, with $y$ as the semantic description of $x$. The collection of semantic descriptions from seen classes by $Y_s$. $c$ is the label of $x$, and the set of labels of seen classes is $C_s$. Furthermore, we have $U_{z}=\{(x, c)|x \in X_u, c \in C_u\} $ and $U_{g}=\{(x, c)|x \in X_s \cup X_u, c \in C_s \cup C_u\} $ as the test sets for ZSL and GZSL respectively, where $X_u$ is a positive sum of visual samples from the unseen classes, and $C_u$ is the set of labels of the unseen classes. The only difference between the ZSL and GZSL is that zero-shot learning adds samples from seen classes into the test set in the testing stage.

	In the language of the above notations, the task of ZSL is to learn a classifier $f_{z}: X_u \rightarrow C_u$ while GZSL is to learn a classifier $f_{g}: X_s \cup X_u \rightarrow C_s \cup C_u$

\subsection{Overview}
	The main idea of this paper is illustrated in Fig.~\ref{fig:f2}. We use GAN to produce samples under the conditional constraints of semantic description.

	The novelty of our method is of two aspects. First, we propose a bi-semantic reconstructing (BSR) component for calculating the reconstruction losses of seen and unseen classes that constrains the generative model. Second, we propose a novel recognition method called visual semantic recognition (VSR) which leverages both visual samples and semantic descriptions to train a classifier, and deploy the BSR component for the reconstruction of the semantic information over test visual samples and concatenate them for the classification. 

	The proposed component and classification method can be flexibly attached to other generative model based ZSL methods to improve their recognition results.

\subsection{Feature Generation}
	Our generative model is built upon GAN~\cite{GAN}, which consist of two components: one discriminator and one generator. The samples of unseen classes for training the classifier are generated by the generator whose inputs are the random noise $z$ sampling from Gaussian distribution and the semantic description $y$ which acts as conditional constraint. Based on these parameters, the generator outputs the synthesized sample of the classes constrained by the semantic description. Afterwards, the discriminator measures the distributional distance between the real samples $x$ and synthesized samples $G(z,y)$, and dscriminate the inputs is real or synthesized samples. The generator is trained to minimize the following loss functions:
	\begin{equation}\label{eq:e1}
		L_G=-\mathbb{E}_{z \sim P_z}[D(G(z,y))]+ \alpha L_{cls} (G(z,y))
	\end{equation}

	where the first term is Wasserstein loss~\cite{WGANGP} , and the second term is the classification loss of synthesized samples from the seen classes , $\alpha$ is the weight of classification loss. Previous methods~\cite{GAZSL, LISGAN} consider the separability between the synthesized samples from seen classes and those from the unseen, and make relative constraints on the generator and the discriminator. However, since there are no unseen visual samples available in the training stage, hastily constraining the classification on the generative model may generate samples that heavily deviate from the real ones. Therefore, we delete those terms for simplification and only constrain the classification for the seen synthesized samples on the generator. As in~\cite{WGANGP}, the discriminator is trained by the loss function below:
	\begin{equation}\label{eq:e2}
		\begin{split}
			L_D=-\mathbb{E}_{z \sim P_z}[D(G(z,y))] -
								 \mathbb{E}_{x \sim P_r}[D(x)] + 
								\\ \beta \mathbb{E}_{\tilde{x} \sim P_{\tilde{x}}}[(\| D(\tilde{x}) \|_2-1)^2]
		\end{split}
	\end{equation}

	where $x$ is the actual visual sample from training set, $\tilde{x}$ is the interpolation of real and synthesized samples. The first two terms supervise the discriminator to output result approximating the Wasserstein distance between the distribution of real samples and that of synthesized samples, and the third term is the gradient penalty which enforces the discriminator to satisfy the Lipschitz constraint and stabilizes the training process. 

\subsection{Bi-semantic component}
	With the help of GAN, we can generate synthesized samples to train a classifier, but this generative model does not guarantee the quality of the synthesized samples. Inspired by~\cite{SAE, DGAN}, we expect to perform the reverse operation of sample generation, which maps the synthesized samples into semantic space, then minimize the reconstruction loss, \emph{i.e.} the least square loss  between the predicted semantic description and the ground truth, in order to improve the quality of data generation. 
	However, since the seen classes and unseen classes belong to two different domains, and the samples is only from the seen classes available in the training stage, the simple use of one regressor for calculating the reconstruction loss may exacerbate the bias of synthesized sample in the unseen classes to the seen classes and therefore decrease the accuracy of GZSL. Therefore, as shown in Fig.~\ref{fig:f2}, we propose a bi-semantic reconstructing (BSR) component which consist of two different regressors $R_s$ and $R_u$ to calculate the reconstruction losses of the seen and unseen to regularize the generator respectively, and optimize themselves.  
	
	The loss of $R_s$ is formulated as follows:
	\begin{equation}\label{eq:e3}
		L_{RS}=\mathbb{E}_{z \sim P_z}\| R_s(G(z,y_s))-y_s \|_2^2
	\end{equation}
	
	Where $y_s$ is the semantic description from seen classes. Similarly, the loss of $R_u$ can be written as:
	\begin{equation}\label{eq:e4}
		L_{RU}=\mathbb{E}_{z \sim P_z}\| R_u(G(z,y_u))-y_u \|_2^2
	\end{equation}
	
	Where $y_u$ is the semantic description from unseen classes. The generator with the two reconstruction losses can improve the relevance between synthesized samples and semantic descriptions and promote the recognition.

	The semantic descriptions from $R_s$ and $R_u$ can be composed by the following simply weighted addition function for the subsequent use:
	\begin{equation}\label{eq:e5}
		\hat{y}=\gamma R_s(x) + (1-\gamma)R_u(x)
	\end{equation}

	Where $\gamma \in [0, 1]$ is the balance parameter, $x$ is the visual sample. In the testing stage, given the test visual sample,we use the reconstructed semantic descriptions $\hat{y}$ from Eq.~\ref{eq:e5} to improve the recognition result.

\subsection{Visual Semantic Recognition}
	When the proposed generative model has been well trained, we can easily get unseen synthesized samples. Therefore, in the zero-shot recognition stage, both the (synthesized) visual samples and semantic descriptions are available for the training of classifier. Most existing methods~\cite{GAZSL, FCLA, DGAN, LISGAN} simply predict labels of test samples by semantic descriptions, or only by visual samples when training a classifier. In contrast, we try to make full use of the available information to train the classifier for a better classification. The visual samples and semantic descriptions are united as the inputs of the classifier to improve the recognition result. In the testing stage, given the test visual samples, the semantic descriptions can be reconstructed by the BSR component from Eq.~\ref{eq:e5}. We call our recognition method as visual semantic recognition (VSR).

	The softmax classifier $P$ is optimized by the following negative log likelihood:
	\begin{equation}\label{eq:e6}
		min_{\psi}\frac{1}{|X|}\sum_{x,y,c \in (X,Y,C)}logP(c|x,y;\psi)
	\end{equation}

	Where $\psi$ is the training parameter, $x$ are the visual samples, $y$ are the semantic descriptions, and $c$ are the labels of $x$.

%------------------------------------------------------------------------
\section{Exiperiments}
\subsection{Dataset}
	We evaluate our method on four benchmark datasets as follows: Caltech-UCSD Birds-200-2011 (CUB)~\cite{CUB}, SUN Attribute (SUN)~\cite{SUN}, APascal-aYahoo (APY) )~\cite{APY}, and Animal with Attributes 2 (AWA2)~\cite{Zongshu1}. 
	
	AWA2 and APY are coarse-gained datasets and with fewer categories. AWA2 contains 37,732 images of 50 types of animals where every animal category is associated with an 85-dimention attribute vector. APY contains 32 categories from both PASCAL VOC 2008 dataset and Yahoo image search engine, which has 15,339 images totally. Every category of objects in APY is annotated by an additional 64-dimensional attribute vector. 
	
	CUB and SUN are fine-grained image datasets. CUB contains 11,788 images of 200 species of birds, which is an extended version to the CUB-200 dataset. Each type of birds is associated with a 312 dimension attribute vector.  SUN is a large-scale scene attribute dataset, which spans 717 categories and 14,340 images in total. Every category is annotated with 102 attribute labels. We report the dataset statistics, the split settings of ZSL and GZSL in Table~\ref{tab:t1}.

	 We extract the visual features from the CNN by 2048-dimensional top-layer pooling units of the ResNet-101~\cite{RESNET} from the real image as our visual samples. ResNet-101 is pre-trained on ImageNet~\cite{ImageNet}. For the semantic descriptions, we utilize the default attributes~\cite{Attribute} on all the four datasets. For fair comparisons, we adopt the splits, classes embedding and evaluation metrics proposed in ~\cite{Zongshu1} for evaluation. 

\begin{table}[t]
\begin{center}
\begin{tabular}{|l|c|c|c|c|}
\hline
Dataset & CUB & SUN & AWA2 & APY \\
\hline
Samples & 11,788 & 14,345 & 37,732 & 15,339 \\
Attrubutes & 312 & 102 & 85 & 64 \\
Seen Classes & 150(50) & 645(65) & 27(13) & 20(5) \\
Unseen Classes & 50 & 72 & 72 & 12 \\
\hline
\end{tabular}
\end{center}
\caption{Dataset statistics. The (number) in Seen Classes is the number of test classes used in test stage of GZSL.}\label{tab:t1}
\end{table}

%zsl
\begin{table}[t]
\begin{center}
\begin{tabular}{|l|c|c|c|c|}
\hline
Dataset & CUB & SUN & AWA2 & APY \\
\hline
DAP~\cite{DAP}&40.0&39.9&46.1&33.8 \\
SSE~\cite{SSE}&43.9&51.5&61.0&34.0 \\
DEVISE~\cite{DEVISE}&52.0&56.5&59.7&39.8 \\
SJE~\cite{SJE}&53.9&53.7&61.9&32.9 \\
ALE~\cite{ALE}&54.9&58.1&62.5&39.7 \\
SAE~\cite{SAE}&33.3&40.3&54.1&8.3 \\
PSR~\cite{PSRZSL}&56.0&61.4&63.8&38.4 \\
GAZSL~\cite{GAZSL}&55.8&61.3&68.4&41.1 \\
F-CLSWGAN~\cite{FCLA}&57.3&60.8&68.8&40.5 \\
DGAN~\cite{DGAN}&51.0&54.8&67.7&40.4 \\
LISGAN~\cite{LISGAN}&58.8&61.7&71.2&43.1\\
\hline
SR(ours)&57.7&61.2&68.4&41.3 \\
BSR(ours)&61.3&62.1&69.9&43.5 \\
BSR+VSR(ours)&\textbf{61.9}&\textbf{64.0}&\textbf{71.6}&\textbf{47.1} \\

\hline
\end{tabular}
\end{center}
\caption{The results of our proposed method and existing methods for ZSL. The best results are highlighted with bold numbers.}\label{tab:t2}
\end{table}

%gzsl
\begin{table*}[t]
\begin{center}
\begin{tabular}{|l|ccc|ccc|ccc|ccc|}
\hline
\multirow{2}{*}{Methods} &\multicolumn{3}{|c|}{CUB}&\multicolumn{3}{|c|}{SUN}&\multicolumn{3}{|c|}{AWA2}&\multicolumn{3}{|c|}{APY}  \\
\cline{2-13}
\multirow{2}{*}{}&\textbf{u}&\textbf{s}&\textbf{h}&\textbf{u}&\textbf{s}&\textbf{h}&\textbf{u}&\textbf{s}&\textbf{h}&\textbf{u}&\textbf{s}&\textbf{h} \\
\hline
DAP~\cite{DAP}&1.7&\textbf{67.9}&3.3&4.2&25.2&7.2&0.0&84.7&0.0&4.8&78.3&9.0 \\
SSE~\cite{SSE}&8.5&46.9&14.4&2.1&36.4&4.0&8.1&82.5&14.8&0.2&78.9&0.4  \\
DEVISE~\cite{DEVISE}&23.8&53.0&32.8&16.9&27.4&20.9&17.1&74.7&27.8&4.9&76.9&9.2  \\
SJE~\cite{SJE}&23.5&53.0&32.8&14.7&30.5&19.8&8.0&7.9&14.4&3.7&55.7&6.9  \\
ALE~\cite{ALE}&23.7&62.8&34.4&21.8&33.1&26.3&14.0&81.8&23.9&4.6&76.9&9.2  \\
SAE~\cite{SAE}&7.8&54.0&13.6&8.8&18.0&11.8&1.1&82.2&2.2&0.4&\textbf{80.9}&0.9 \\
PSR~\cite{PSRZSL}&24.6&54.3&33.9&20.8&37.2&26.7&20.7&73.8&32.3&13.5&51.4&21.4  \\
GAZSL~\cite{GAZSL}&23.9&60.6&34.3&21.7&34.5&26.7&19.2&\textbf{86.5}&31.4&14.6&78.2&24.0  \\
F-CLSWGAN~\cite{FCLA}&43.7&57.7&49.7&42.6&36.6&39.4&54.1&67.9&60.2&32.9&61.7&42.9  \\
DGAN~\cite{DGAN}&39.3&66.7&49.5&35.2&24.7&29.1&32.1&67.5&43.5&30.4&75.0&43.4  \\
LISGAN~\cite{LISGAN}&46.5&57.9&51.6&42.9&37.8&40.2&44.9&77.5&56.9&\textbf{34.3}&68.2&45.7 \\
\hline
SR(ours)&41.8&60.7&49.5&\textbf{47.2}&34.5&39.8&52.5&71.3&60.4&30.8&71.4& 43.0 \\
BSR(ours)&46.4&56.7&51.0&41.4&\textbf{39.5}&40.4&55.9&69.3&61.9&33.8&57.0& 42.4 \\
BSR+VSR(ours)&\textbf{48.3}&60.1&\textbf{53.5}&45.8&38.1&\textbf{41.6}&\textbf{58.0}&68.9&\textbf{63.0}&34.1&70.7&\textbf{46.0}  \\

\hline
\end{tabular}
\end{center}
\caption{The results of our proposed method and existing methods for GZSL. The best results are highlighted with bold numbers.}\label{tab:t3}
\end{table*}

\subsection{Evaluation}
 	 Following the previous work ~\cite{Zongshu1, LISGAN}, for ZSL, we report the average per-class top-1 accuracy \textbf{a} for each evaluated method. For GZSL, with the same setting in ~\cite{Zongshu1}, we calculate average per-class top-1 accuracy on seen classes denoted as \textbf{s}, and the average per-class top-1 accuracy on unseen classes denoted as \textbf{u}, and their harmonic mean $\textbf{h}=2*(\textbf{s}*\textbf{u})/(\textbf{s}+\textbf{u})$ 

\subsection{Implementation details}
	We apply the generator, discriminator, classifier, and the two regressors as MLPs, which have one or two hidden layers with ReLU activation. The last layer of MLPs of all components in our model is a linear layer. The weight of gradient penalty $\beta$ is set to 10, which follows the setting of ~\cite{WGANGP}. Since we use the numbers of synthesized samples to control the reconstruction losses, both weights of reconstruction loss $L_{RS}$ and $L_{RU}$ are simply set to 1. The noise $z$ is samples from Gaussian distribution conditioned by the semantic description $y$, severing as the inputs of Generator. The compared methods are representative and state-of-the-art ones published in the last few years, which are reported very recently. Specially, we compared our methods with DAP~\cite{DAP}, SSE~\cite{SSE}, ALE~\cite{ALE}, DEVISE~\cite{DEVISE}, SJE~\cite{SJE}, SAE~\cite{SAE}, PSR~\cite{PSRZSL}, F-CLSWGAN~\cite{FCLA}, GAZSL~\cite{GAZSL}, DGAN~\cite{DGAN}, LISGAN~\cite{LISGAN}.

\subsection{Zero-shot Learning}
	We compared our result with recent state of art zero-shot learning methods in Table~\ref{tab:t2} on four datasets. In these experiments, the labels of test samples only belong to $C_u$ disjoint with the training label set $C_s$. 

	For the lack of test results on AWA2, we recreate the GAZSL~\cite{GAZSL}, F-CLSWGAN~\cite{FCLA}, LISGAN~\cite{LISGAN} to test the recognition results on AWA2, whose code is available online. We use the same classifier as its setting on other datasets. The results of these three methods on other datasets are copy from ~\cite{LISGAN}. The GDAN method is close to our model. However, it has no ZSL test results. In order to make a good comparison, we recreate GDAN. Following the setting of its original paper, we use the nearest neighbor prediction to predict the recognition results. The result of other methods is copy from ~\cite{Zongshu1}. 

	We perform additional analyses to gain the further insight about the methods we proposed. In order to show the ability of BSR component, we train two different generative models to make fully comparison. The first one only applies one semantic reconstructing regressor to reconstruct both seen and unseen reconstruction loss, which we define as SR. The second one applies the BSR component we proposed to reconstruct the seen and unseen reconstruction loss by two different regressors. In the setting of the two methods, we simply deploy a softmax classifier to make a fair comparison. Unsurprisingly, the result on zero-shot learning show that using BSR component could make a better perform than using one regressor simply to improve the quality of generated data. Meanwhile, we show the improvements of the result by BSR component with the help of the proposed VSR method. It is found that the BSR + VSR can achieve a further improvements. We can see that this method achieves the best on all these datasets. Specially, our method has achieved up to 3.1\%, 2.3\%, 0.4\%, and 4.0\% improvements on CUB, SUN, AWA2, and APY.

%column
\begin{figure*}
\begin{center}
%\fbox{\rule{0pt}{2in} \rule{.9\linewidth}{0pt}}
	\subfigure[CUB]{
		\centering
		\includegraphics[width=0.3\linewidth]{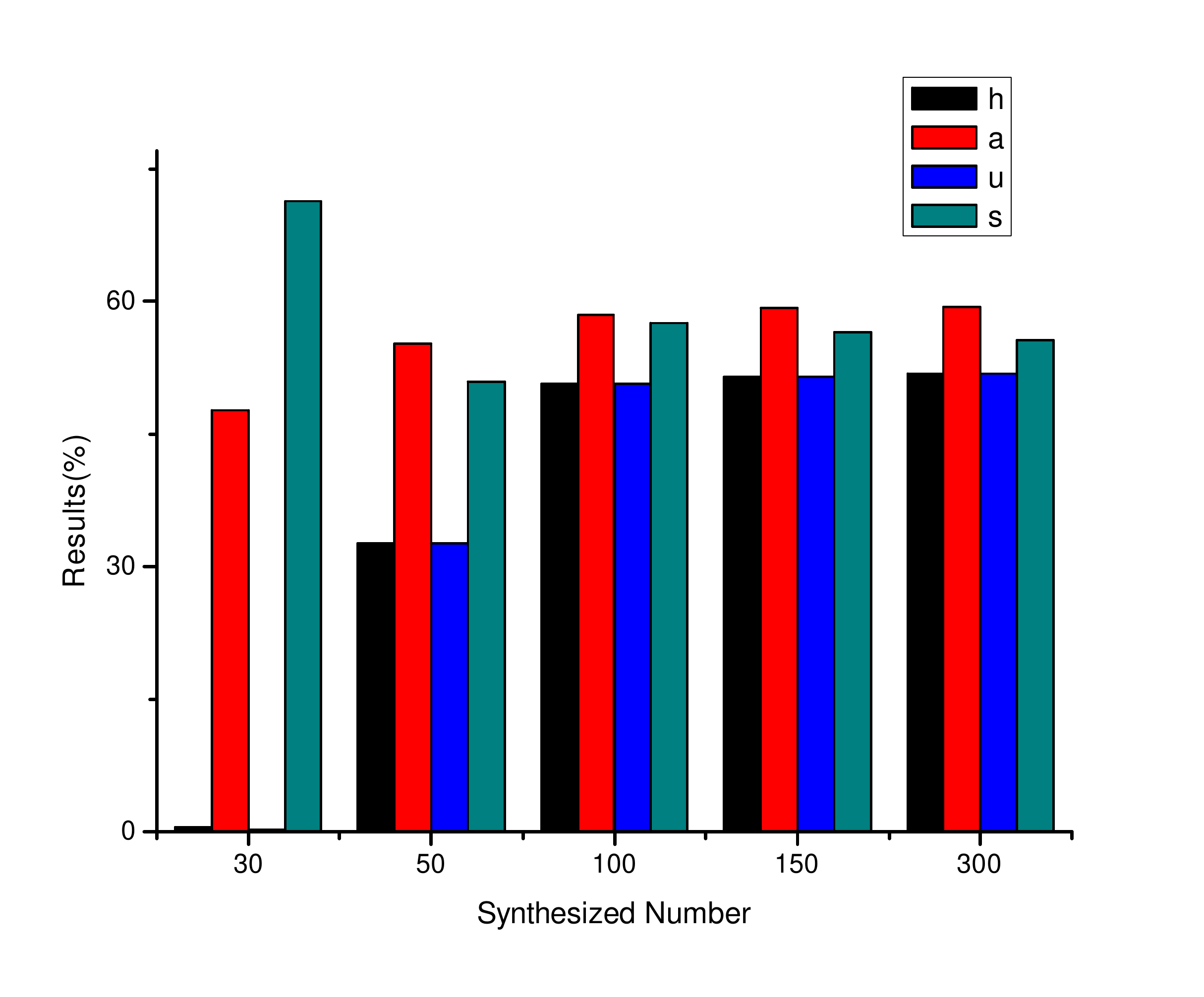}
	}
	\subfigure[SUN]{
		\centering
		\includegraphics[width=0.3\linewidth]{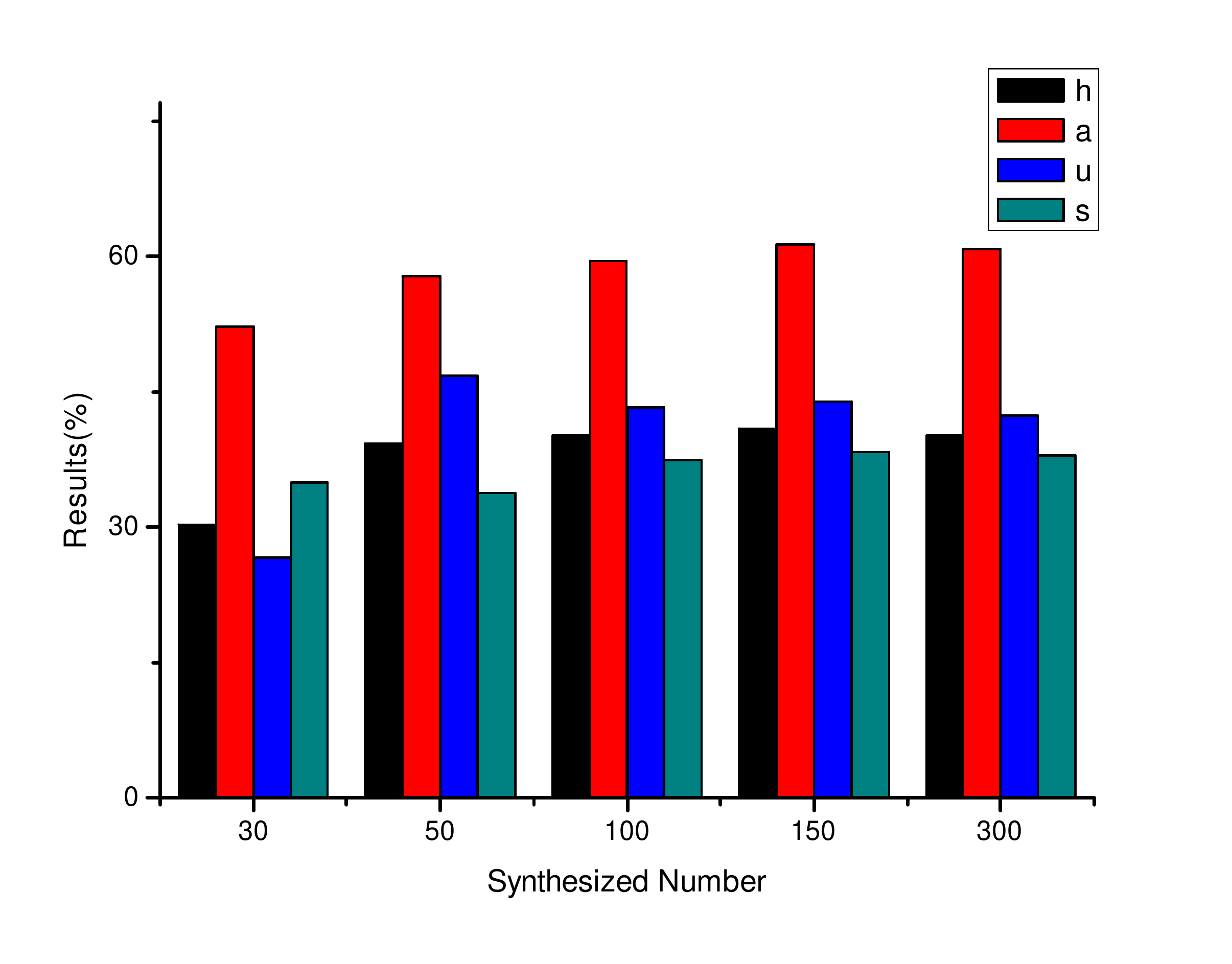}
	}
	\subfigure[AWA2(Small Scale)]{
		\centering
		\includegraphics[width=0.3\linewidth]{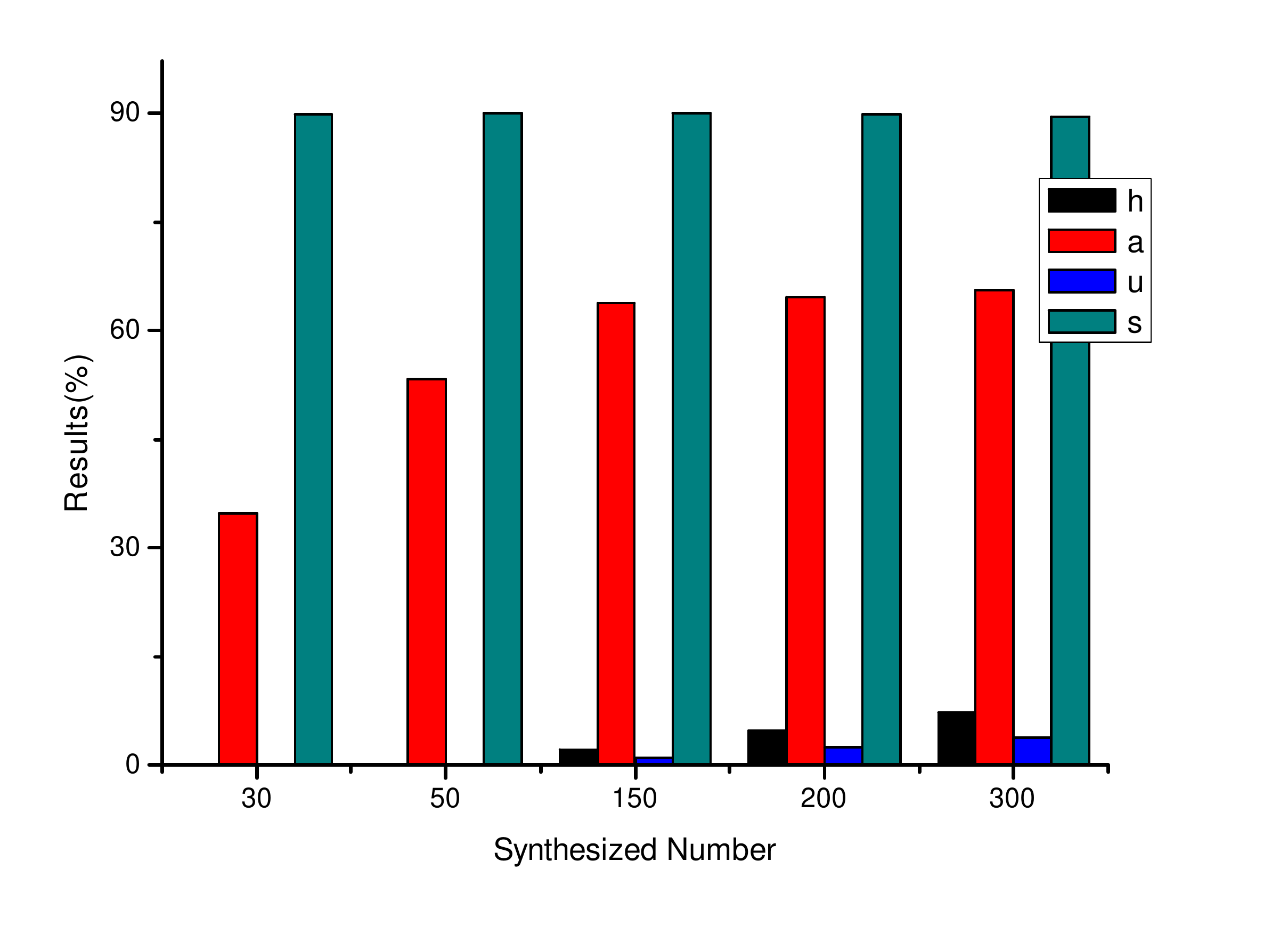}
	}
	\subfigure[APY(Small Scale)]{
		\centering
		\includegraphics[width=0.3\linewidth]{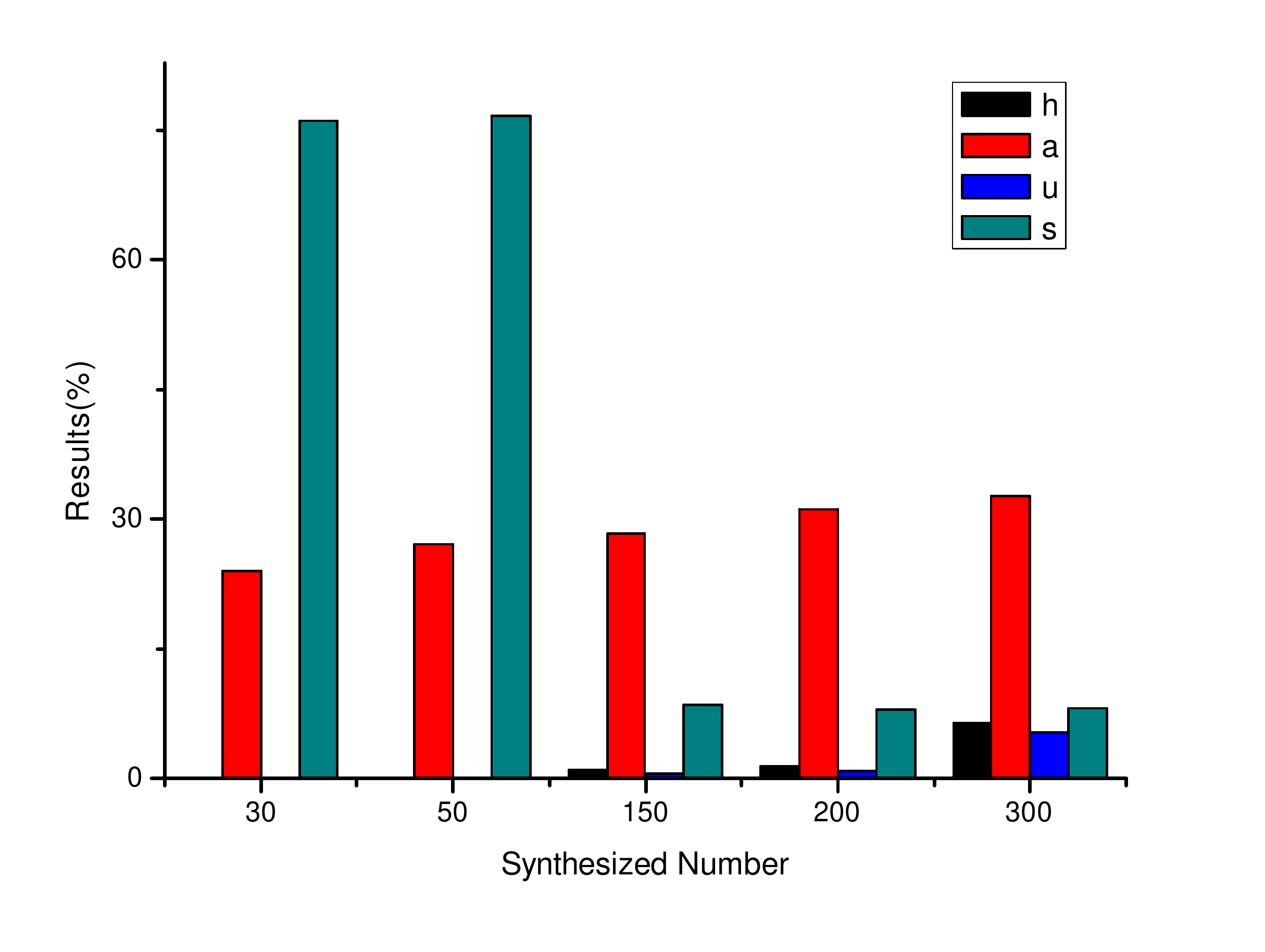}
	}
	\subfigure[AWA2(Large Scale)]{
		\centering
		\includegraphics[width=0.3\linewidth]{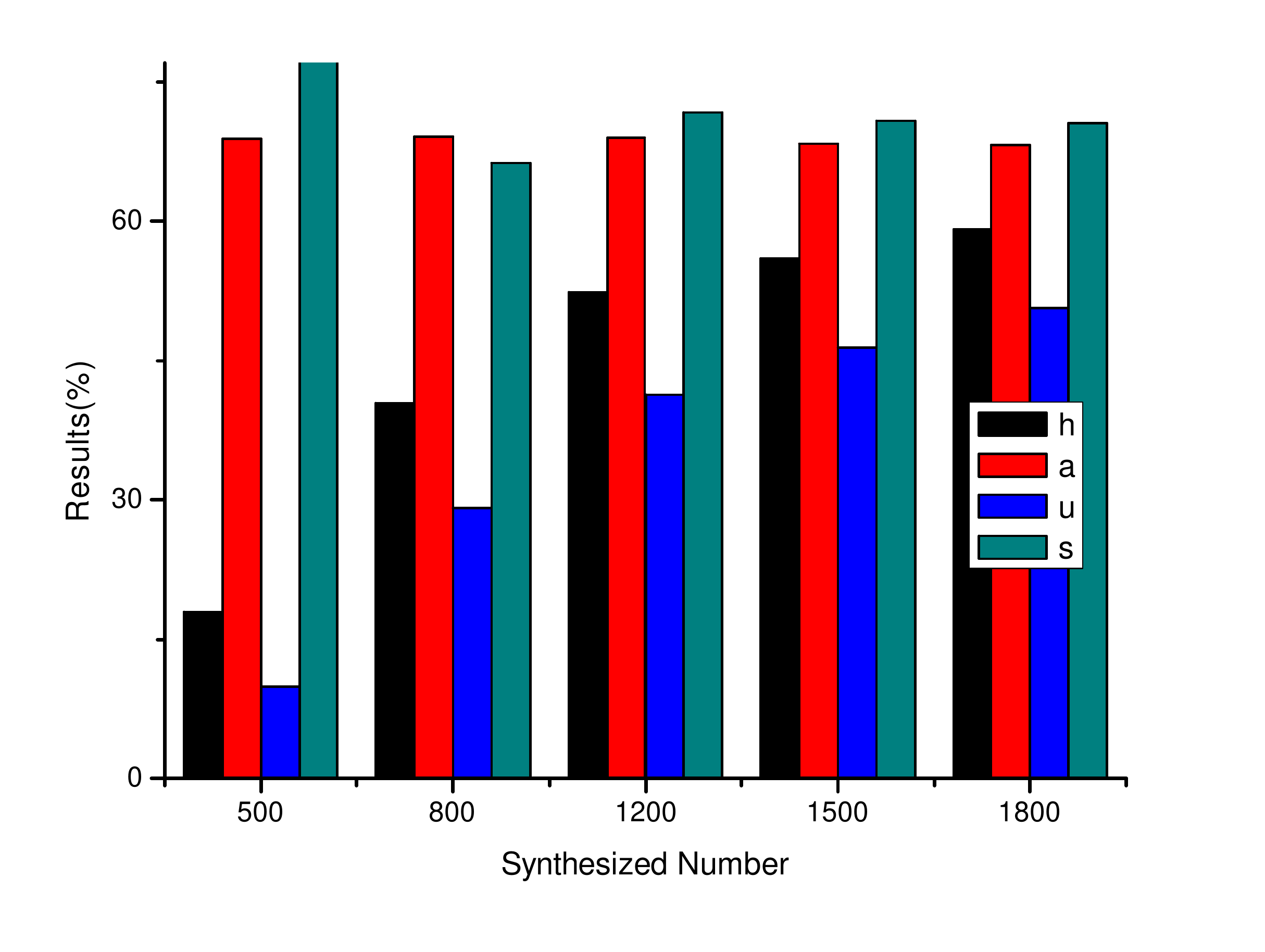}
	}
	\subfigure[APY(Large Scale)]{
		\centering
		\includegraphics[width=0.3\linewidth]{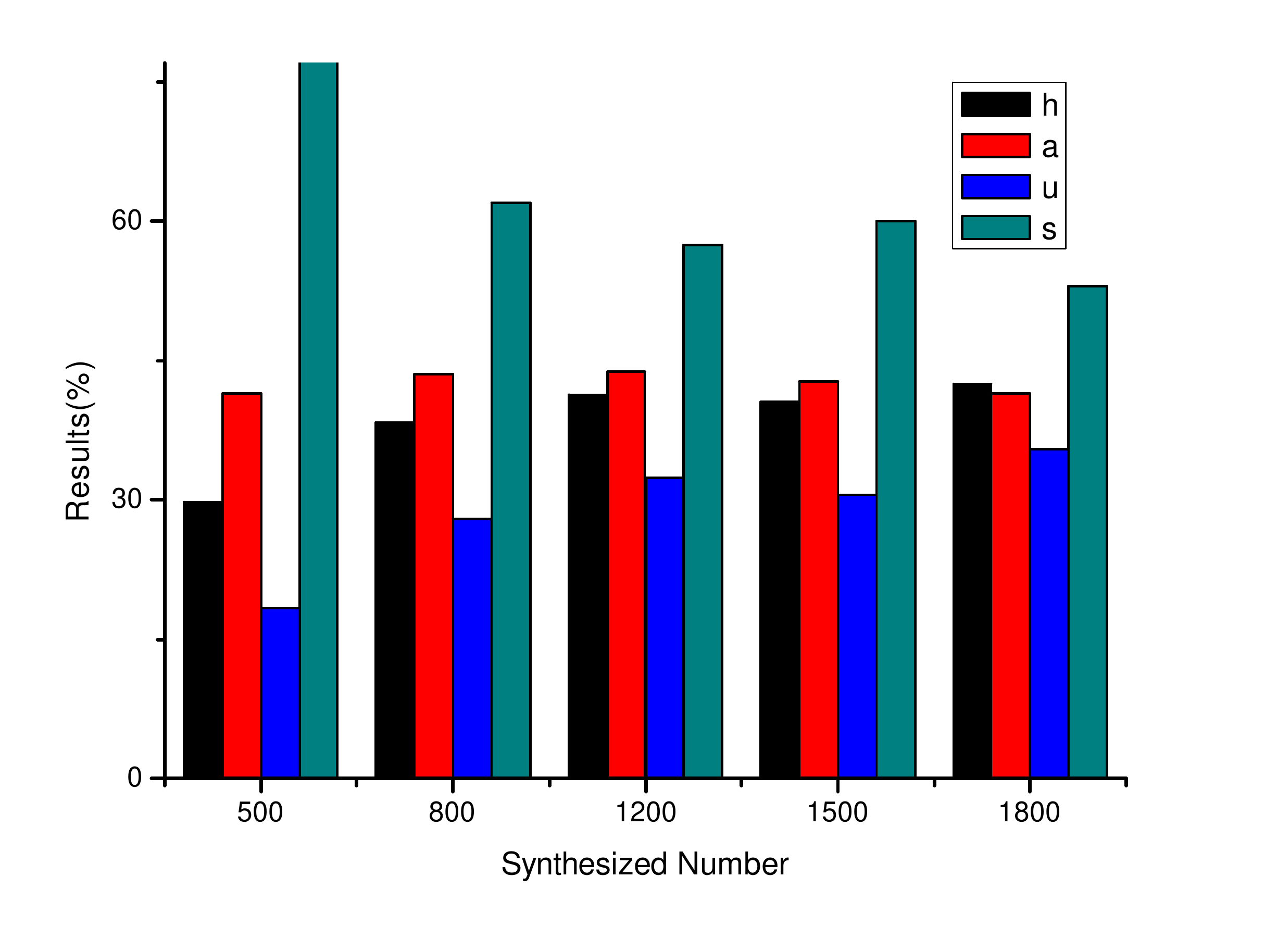}
	}

\end{center}
   \caption{The affection of synthesized numbers on different datasets. \textbf{a} is the score for ZSL. \textbf{s}, \textbf{u}, and \textbf{h} is the score for GZSL}
\label{fig:f3}
\end{figure*}

%stable
\begin{figure}[t]
\begin{center}
	\subfigure[ZSL]{
		\centering
		\includegraphics[width=0.4\linewidth]{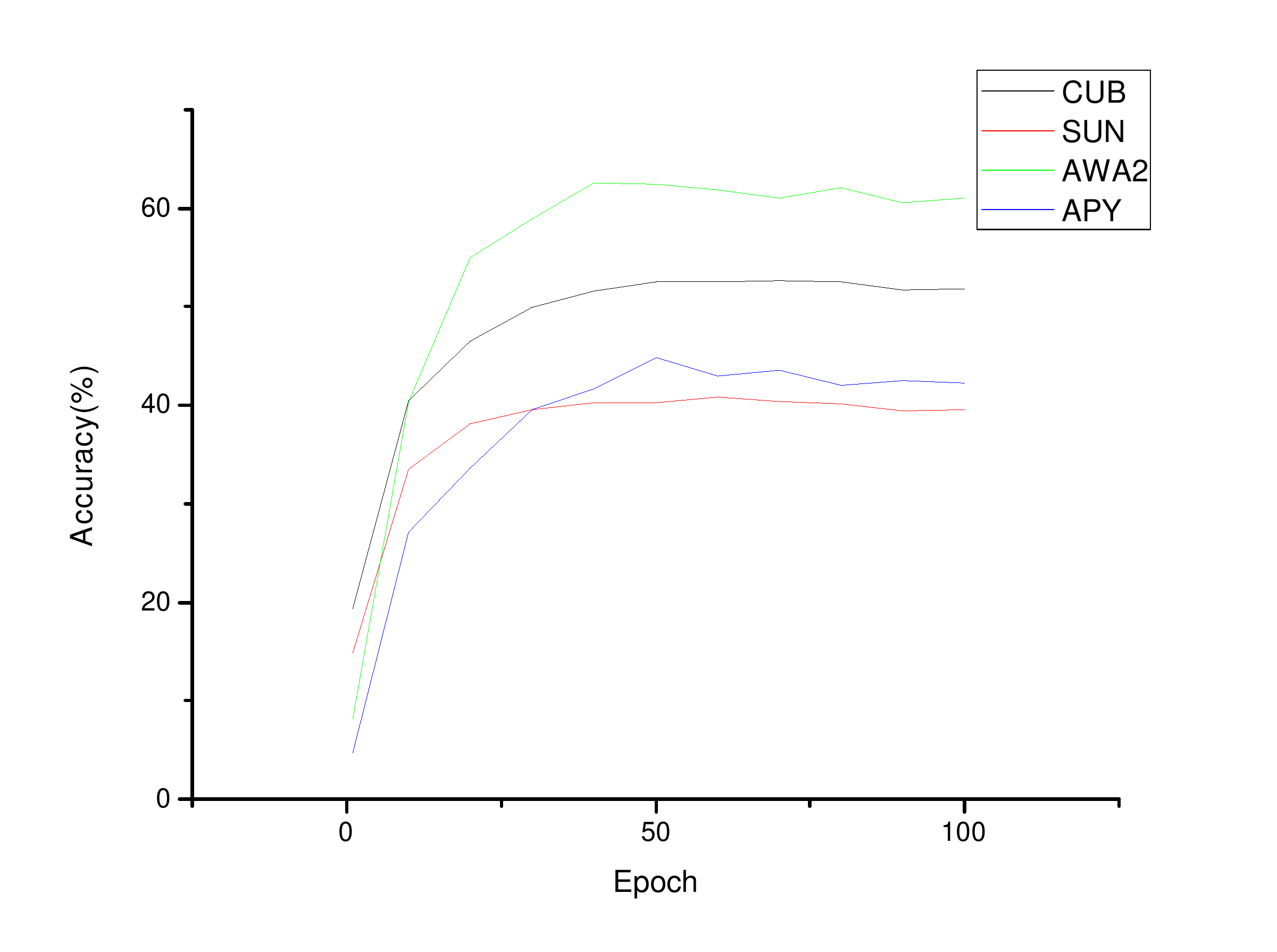}
	}
	\subfigure[GZSL]{
		\centering
		\includegraphics[width=0.4\linewidth]{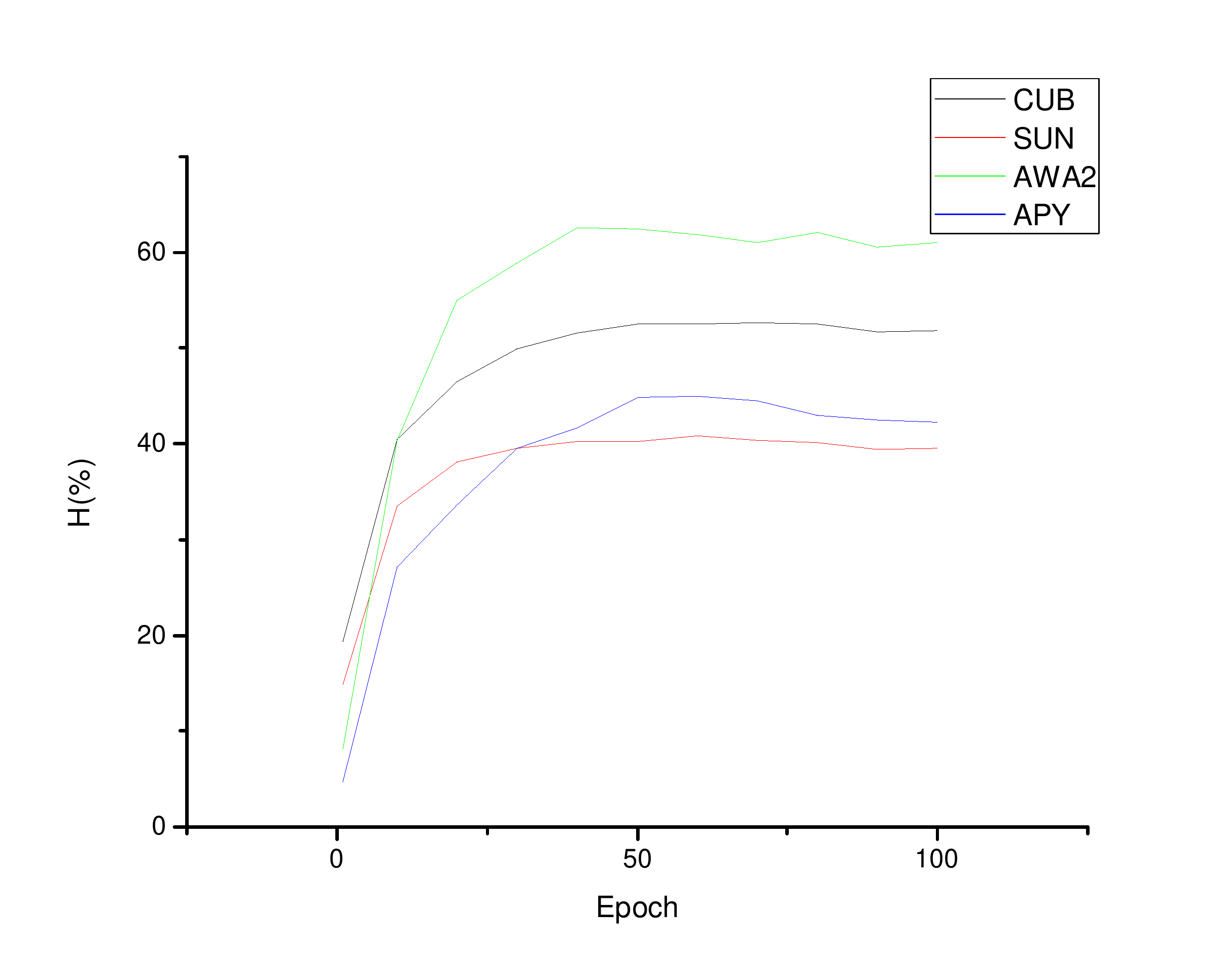}
	}

\end{center}
   \caption{The result curve on different test set changes with the increase of training epoch. }
\label{fig:f4}
\end{figure}

%gama
\begin{figure*}
\begin{center}
%\fbox{\rule{0pt}{2in} \rule{.9\linewidth}{0pt}}
	\subfigure[CUB]{
		\centering
		\includegraphics[width=0.23\linewidth]{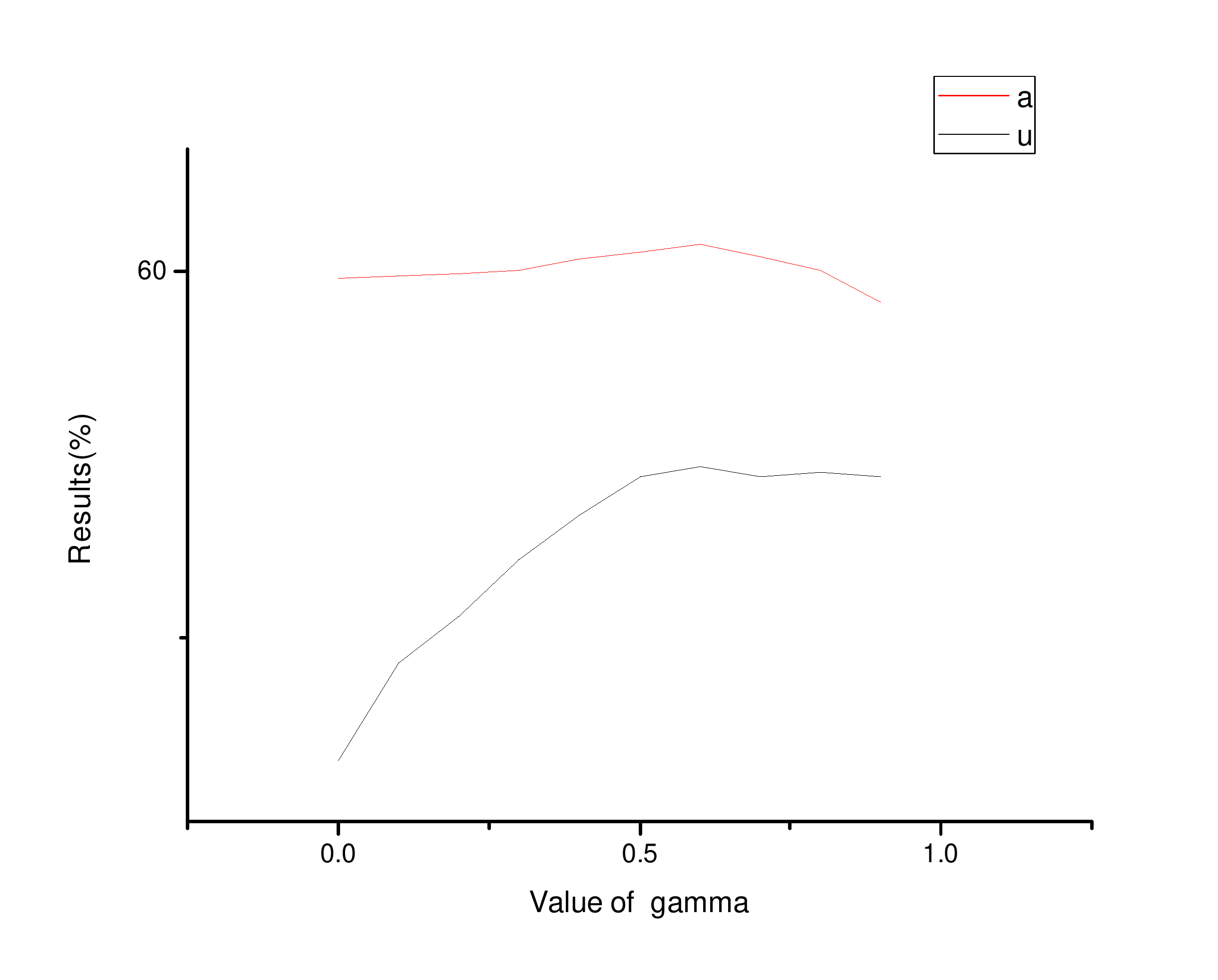}
	}
	\subfigure[SUN]{
		\centering
		\includegraphics[width=0.23\linewidth]{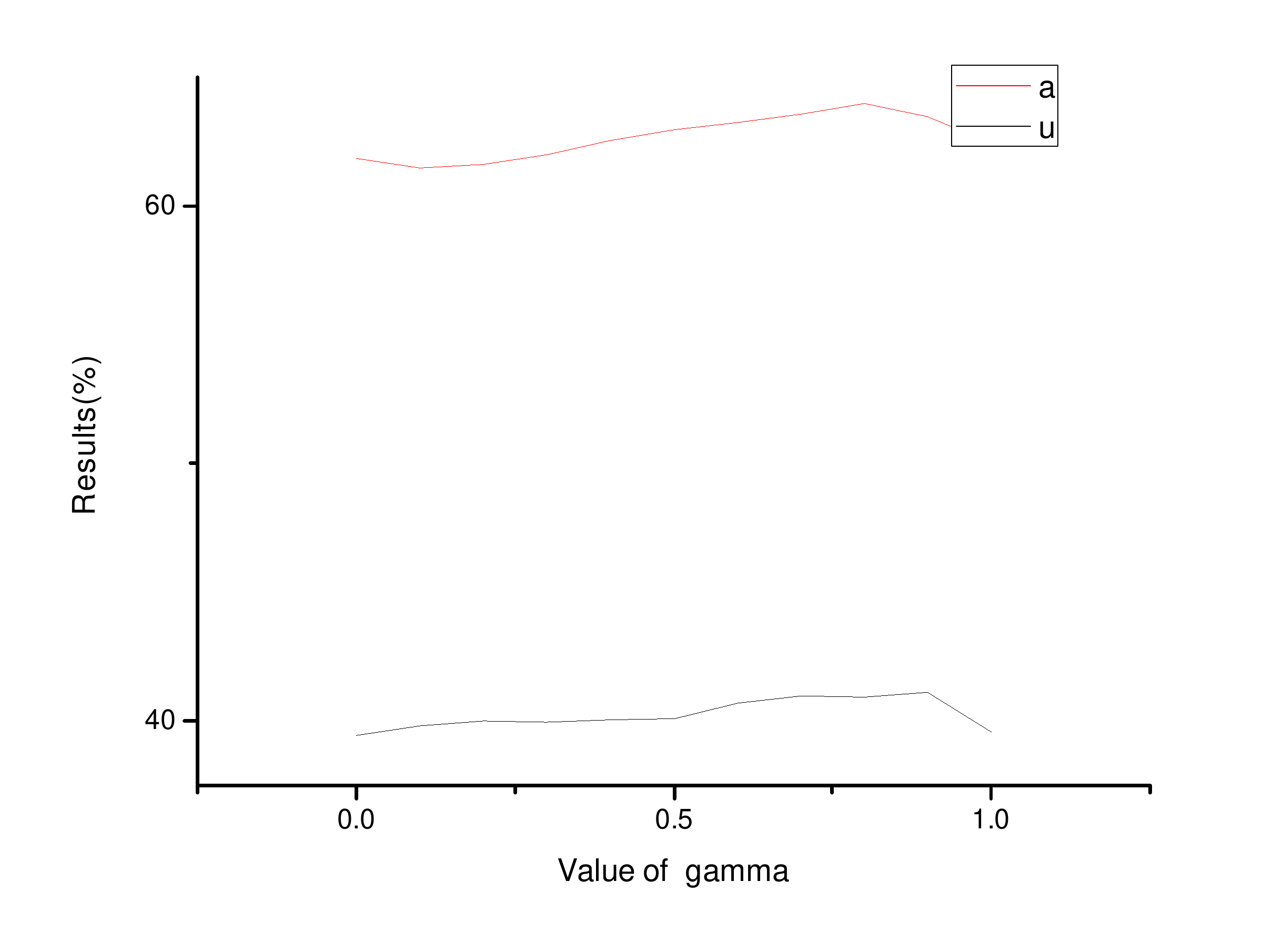}
	}
	\subfigure[AWA2]{
		\centering
		\includegraphics[width=0.23\linewidth]{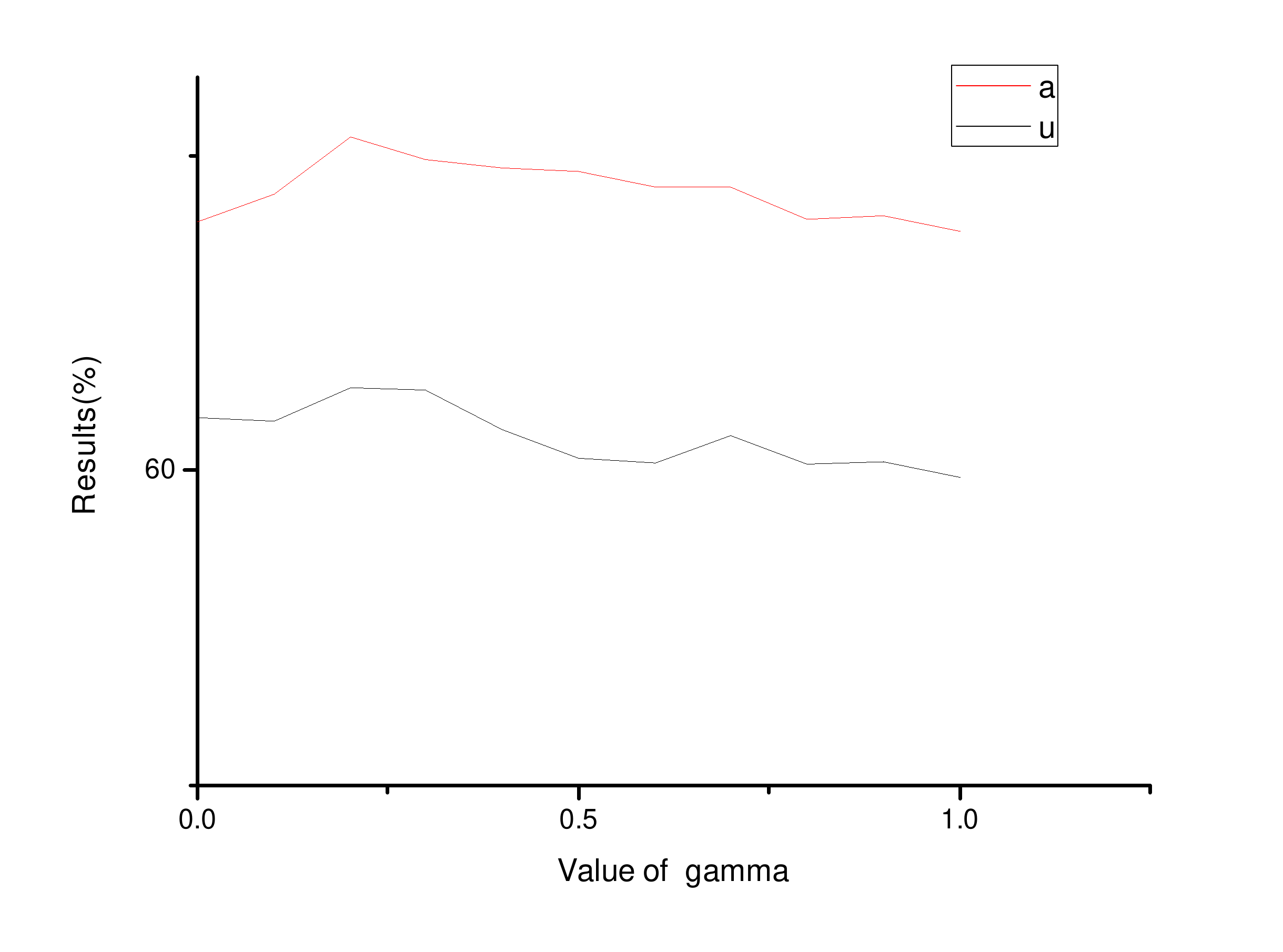}
	}
	\subfigure[APY]{
		\centering
		\includegraphics[width=0.23\linewidth]{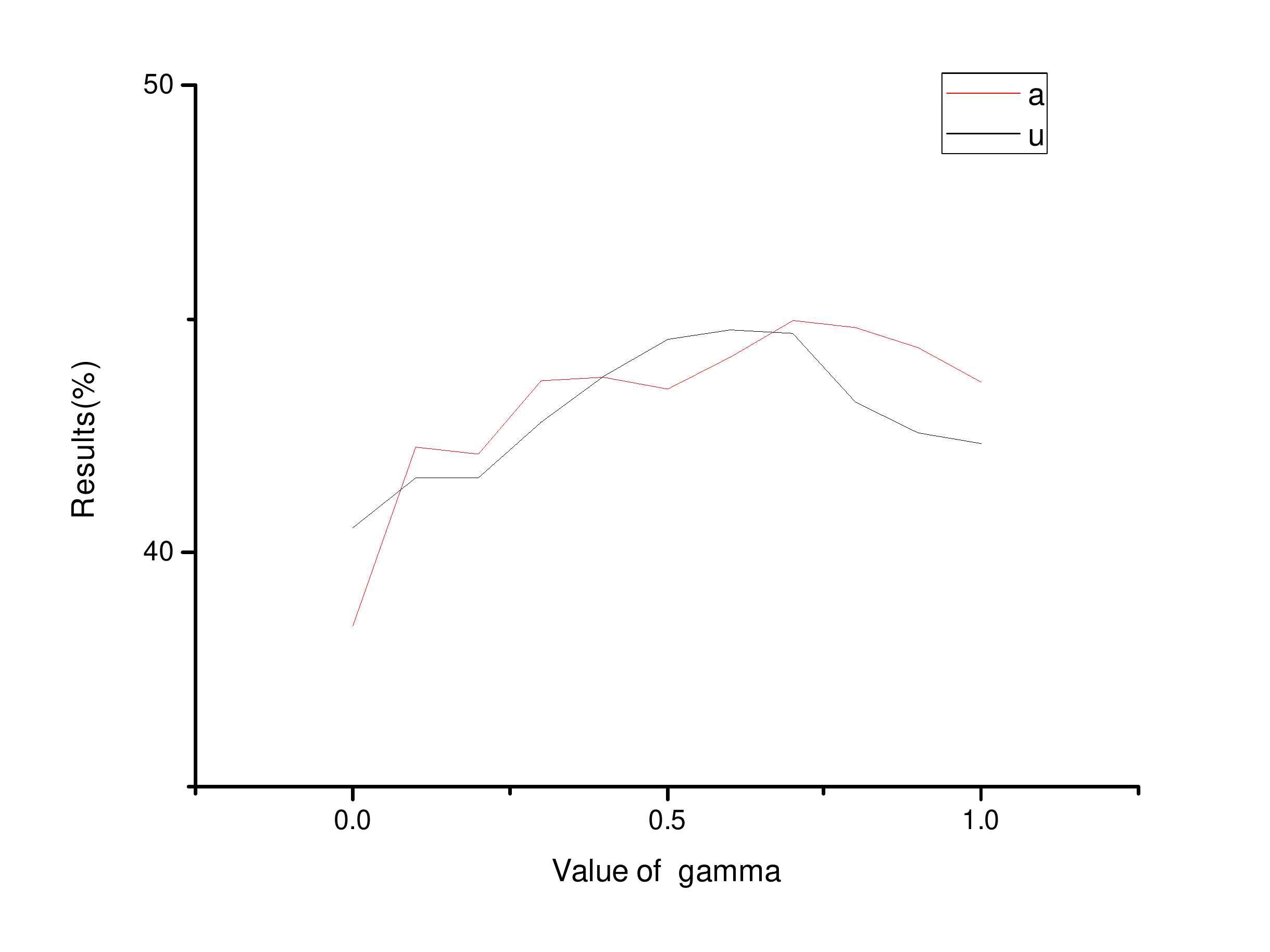}
	}
\end{center}
   \caption{The effect of $\gamma$. The best value of $\gamma$ is always around the ratio between numbers of seen classes and numbers of all classes. }
\label{fig:f5}
\end{figure*}

%as
\begin{figure}[t]
\begin{center}
	\includegraphics[width=0.8\linewidth]{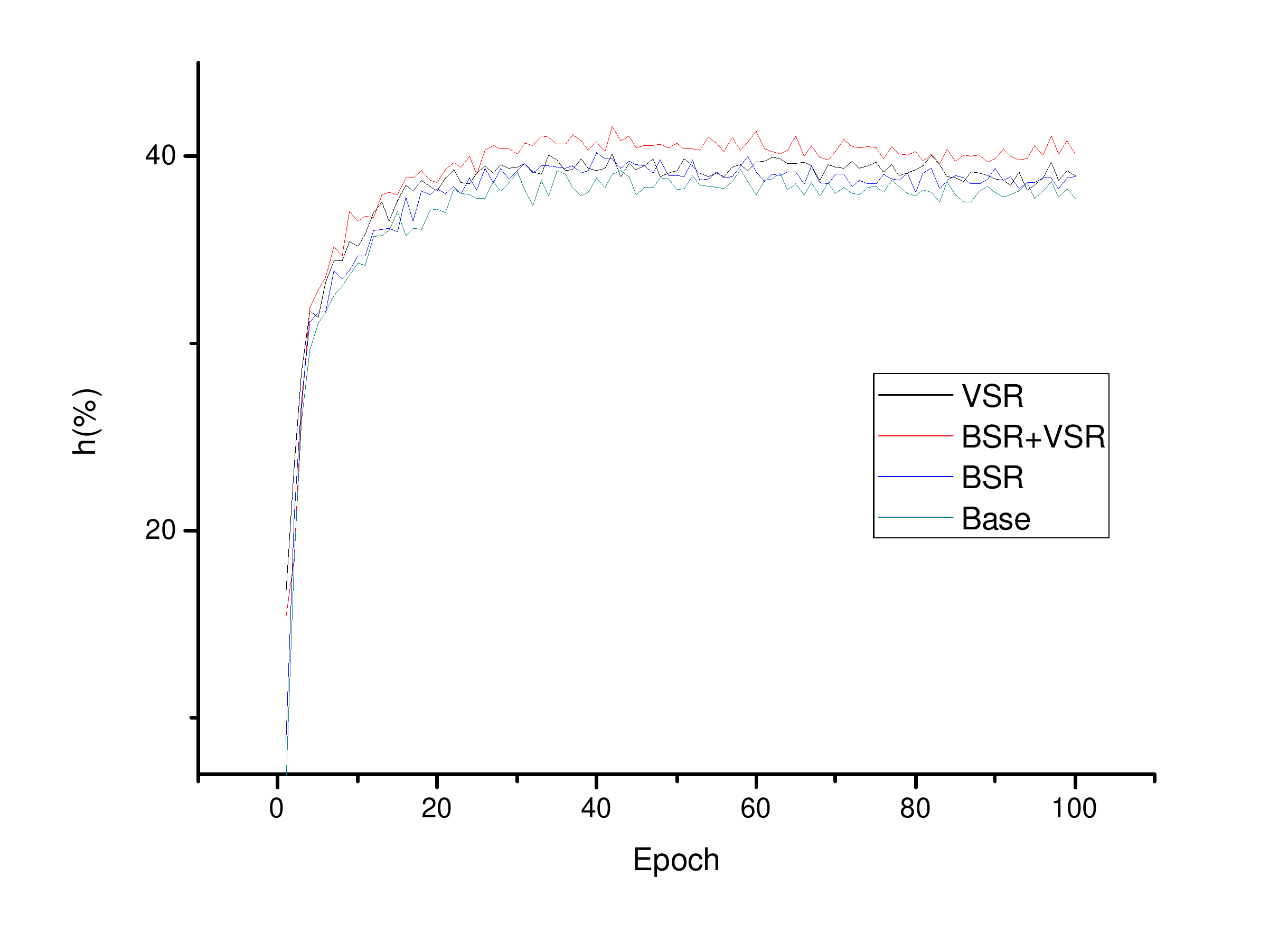}
\end{center}
   \caption{\textbf{u} score of test samples from SUN dataset have different settings: base, BSR, VSR, and BSR+VSR.}
\label{fig:f6}
\end{figure}

\subsection{Generalized Zero-shot Learning}
	The results of our methods of generative zero-shot learning are show on Table~\ref{tab:t3}. The setting at the training stage of the GZSL is the same as ZSL. At the test stage, the test samples of GZSL come from the union of seen and unseen set. Similar to zero-shot learning, we make extra test for GAZSL~\cite{GAZSL}, F-CLSWGAN~\cite{FCLA}, LISGAN~\cite{LISGAN} on AWA2. For the numerical anomaly of GDAN~\cite{DGAN} on the \textbf{s} score of SUN, we retested the results of this method on the dataset, where the classification method is the nearest neighbor prediction following the original setting of the method. 
	
	Similar to the ZSL, we test three types of generative model. As show on the table, our methods and other generative methods perform evidently better than directly methods. In addition, in contrast to the result of SR and BSR, it is easy to find that SR is easy to suffer the unbalance of the results of \textbf{s} and \textbf{u}, resulting in the lower performance of \textbf{h} at the test stage. This phenomenon can be explained by the fact that there are no seen samples existing at the training stage. It would cause the phenomenon that a single semantic reconstructing regressor will confuse the seen and unseen samples, then make a bad reconstruction of semantic descriptions, and lead an error constrain on generative model. 

	 From the result of Table~\ref{tab:t3}, comparing the BSR and BSR + VSR, we can find that with the help of VSR, we can achieve the improvements both on the \textbf{s} and \textbf{u}, leading to an improvement of \textbf{h}. These improvements mainly come from the additional information of the undamaged semantic descriptions at the training stage of the classifier. Since the semantic descriptions at the test stage is reconstructed but not real, the VSR method is limited by the quality of the reconstructed semantic descriptions. In particular, in the terms of \textbf{h}, our methods has achieved up to 1.9\%, 1.4\%, 2.8\%, and 0.3\% improvements on CUB, SUN, AWA2, and APY, respectively.

\subsection{Stability}
	Fig.~\ref{fig:f4} shows the variation curve about the results of ZSL and GZSL with respect to the number of epochs. We test our method (BSR+VSR). For ZSL, we use Top-1 accuracy \textbf{a}; for GZSL, we use \textbf{h} score. The score is recorded every 10 epoches. The experiment shows that our method has a higher stability on CUB and SUN than on APY and AWA2. Different with the two coarse-grain datasets, both of the SUN and CUB are fine-grain datasets with more stable correlation between visual samples and corresponding semantic descriptions. Therefore, our generative models on these datasets can obtain a more stable mapping relation between the distribution of noise conditioned by semantic descriptions and the distribution of real visual samples, leading to a better performance.

\subsection{Effect of Synthesized Number}
	Having verified the effectiveness of our method, we begin to analyze the effect of hyper-parameters on ZSL and GZSL. At first, we evaluate how the number of the synthetic samples affects. For all of the datasets, we test the top-1 accuracy \textbf{a} for ZSL and the merit of GZSL \textbf{s}, \textbf{u}, and \textbf{h} in Fig.~\ref{fig:f3}. For all of the benchmark datasets, we create five datasets in different sizes with our generative model by the sample for each class on $\{30,50,100,150,300\}$. For AWA2 and APY, we make an extra test on $\{500,800,1200,1500,1800\}$ for comparison.

	In Fig.~\ref{fig:f3}, the result of ZSL has increased rapidly when the synthesized number is small, and remains basically unchanged when the number is large. It is not surprising, because softmax classifier needs enough samples to be well trained.

	As shown in Fig.~\ref{fig:f3} (a), (b), and (c), \textbf{u} and the \textbf{h} score has a rapid increase when the number of the synthesized sample increases. However, \textbf{s} has a decrease. It lies in two causes as follows: (1) At the beginning, the classifier tends to the seen classes. With the increased number of the synthesized samples, this tendency has gone. (2) With the increase of the number of the synthesized samples, the BSR component can obtain more samples, and then get an improvement on the ability to map the visual samples into semantic space. In this way, the reconstruction loss by the BSR component can lead to a better constraint on the generative model.

\subsection{Hyperparameter}
	As shown in Fig.\ref{fig:f5}, we evaluated the effect of different sets of $\gamma$ to our method (BSR+VSR) on different datasets, and find it interesting that the best value of $\gamma$ is always around the ratio of numbers of visible classes to the total number of classes. The ratio value of CUB, SUN, AWA2, and APY is 0.75, 0.90, 0.27, and 0.62, respectively.

	As the Eq.~\ref{eq:e4} shows, in fact, $\gamma \in [0,1]$ is the weight of reconstructed semantic description from $R_s$, and $(1-\gamma)$  is the weight of reconstructed semantic description from $R_u$. They make up the semantic descriptions of VSR at the test stage. The higher the value of gamma, the higher  the message of reconstructed semantic description that biases to seen classes, and decrease of message that biases to unseen classes, therefore, the best value of $\gamma$ is related to the ratio of numbers of seen classes to numbers of all classes.

\subsection{Ablation Study}
	In order to gain a further insight whether the BSR and VSR can help us achieve a better performance, we test \textbf{u} score under four settings on SUN for each epoch: base, BSR, VSR and BSR+VSR. A generative model optimized by Eq.~\ref{eq:e1} and Eq.~\ref{eq:e2} is used as base setting. BSR setting is based on the base setting, and the BSR component is utilized to calculate the reconstruction loss to constraint the synthesized samples from generative model. Both the base setting and BSR setting train a simply softmax classifier for test. In VSR setting, the loss function of generator has no reconstruction regularization term, but only the BSR component was trained to mapping the visual samples into semantic space.  In addition, the input of classifier in the test has combined the visual samples and its semantic descriptions. VSR+BSR is the comprehensive method of the BSR setting and VSR setting.
	
	The results show in Fig.~\ref{fig:f6}. Compared  with the base setting, result of the setting  with only BSR or VSR makes a little improvements and the BSR+VSR got the best result. These results verify the effectiveness of BSR and VSR, and the progress made by combining them.

\section{Conclusion}
	This paper proposes a novel zero-shot learning method by taking the advantage of generative adversarial network. Specially, we proposed a bi-semantic reconstructing component with two semantic regressors to reconstruct the seen and unseen synthesized samples into semantic descriptions respectively and calculate the reconstruction losses to constraint the generative model. At the zero-shot recognition stage, we propose a visual semantic recognition method to improve the recognition accuracy. Experiments show that our method outperforms the existing methods with the help of bi-semantic reconstructing component and visual semantic recognition method.

%-------------------------------------------------------------------------

{\small
\bibliographystyle{ieee_fullname}
\bibliography{egbib}
}

\end{document}